# What you get is not always what you see—pitfalls in solar array assessment using overhead imagery

*Wei Hu[1], Kyle Bradbury[1], Jordan M. Malof[2], Boning Li[2], Bohao Huang[2], Artem Streltsov[1], K. Sydny Fujita[3], and Ben Hoen[3]*

[1]Energy Initiative, Duke University
[2]Department of Electrical & Computer Engineering, Duke University
[3]Lawrence Berkeley National Laboratory

***Abstract*—Effective integration planning for small, distributed solar photovoltaic (PV) arrays into electric power grids requires access to high quality data: the location and power capacity of individual solar PV arrays. Unfortunately, national databases of small-scale solar PV do not exist; those that do are limited in their spatial resolution, typically aggregated up to state or national levels. While several promising approaches for solar PV detection have been published, strategies for evaluating the performance of these models are often highly heterogeneous from study to study. The resulting comparison of these methods for practical applications for energy assessments becomes challenging and may imply that the reported performance evaluations overly optimistic. The heterogeneity comes in many forms, each of which we explore in this work: the level of spatial aggregation (e.g. array-level vs regional estimates), the validation of ground truth (manual annotation of imagery vs known solar PV locations), inconsistencies in the training and validation datasets (e.g. different datasets are used for each study and those data are not always made accessible), and the degree of diversity of the locations and sensors (e.g. different satellites, aerial photography ) from which the training and validation data originate. For each, we discuss emerging practices from the literature to address them or suggest directions of future research. As part of our investigation, we evaluate solar PV identification performance in two large regions: the entire state of Connecticut and the city of San Diego, CA. In Connecticut, we also use 33,114 known parcel-level solar PV installations from Berkeley Lab's Tracking the Sun dataset to evaluate parcel-level performance and evaluate capacity estimates using 169 municipalities. We also make our code (which we call *SolarMapper*), pre-trained models, training data, and predictions publicly available and provide a web portal for interactively inspecting each prediction that was made. Our findings suggest that traditional performance evaluation of the automated identification of solar PV from satellite imagery may be optimistic due to common limitations in the validation process. The takeaways from this work are intended to inform and catalyze the large-scale practical application of automated solar PV assessment techniques by energy researchers and professionals.**

## I. Introduction

The quantity of solar photovoltaic (PV) arrays has grown rapidly in the United States in recent years [1,2], with a large proportion of this growth coming from small-scale, or distributed, PV arrays [3]. These systems are less than 1 MW in rated capacity and typically 5-10 kW for household installation. Distributed solar PV offers many benefits [4], but integrating distributed solar PV into existing electric power grids remains challenging due to the intermittency of its generation. A key ingredient for understanding PV growth factors, conducting planning activities that include effectively integrating new PV into existing electric power grids, and encouraging commercial and academic research innovation is highly accurate, high spatial-resolution data on solar PV. This includes, for example, the location, size, and power generating capacity of existing arrays.

### A. Limitations in Solar PV data collection methodologies

The most straightforward approach to gathering small-scale solar PV data would be to curate it from existing sources; however, except for rare exceptions, such data do not exist at scale. Several organizations have begun collecting or publishing PV information, including the Solar Energy Industries Association [5], and government agencies, such as the US Energy Information Administration (EIA)[3,6]. However, existing sources are often limited in spatial resolution or are difficult to obtain, such as those that are proprietarily held. Existing methods of obtaining data on distributed solar PV, such as surveys and utility interconnection filings, are often either unavailable publicly or time consuming to retrieve through documents that have yet to be fully digitized. Additionally, these resources are sometimes incomplete, or heterogeneous throughout a country, or even a smaller political boundary. When they are available, these data are also typically limited in spatial resolution to the state or national level [3]. For example, the EIA began reporting state-level distributed PV data at the end of 2015 [6], which was at the time and remains for the U.S. the highest spatial resolution data on distributed solar PV. No comprehensive database of small-scale solar PV exists in the United States or, as of this writing, for any other country in the world, save one: the United Kingdom [7].

The U.K. database, the most comprehensive to-date, utilized a crowdsourcing campaign and has arguably been the most successful methodology applied at-scale for distributed solar PV data collection. However, drawback of this methodology may be difficulty in scaling it up globally. The database from Stowell et al. [7] includes an estimated 86% of the installed solar PV capacity from the U.K. from the efforts of 343 Open Street Map (OSM) community members (with 95% of contributions from just 11 exceedingly active members). The authors note that the agreement between the area of the OSM-user annotated polygons and the capacity of the system achieved an $R^2$ value of 0.59 due to the fact that the polygon area may cover the "whole extent of a solar farm or installation, not just the PV panel surface," which indicates some heterogeneity in the annotation process – common for manual annotations from multiple annotators. Beyond this project, the OSM community has been exceedingly responsive to acute humanitarian needs for crowdsourced mapping projects in the past [8], although typically covering smaller regions. Given the manual effort required, however, this approach may be genuinely difficult to scale up globally or to country-wide assessments for large nations, and to keep up-to-date.

If distributed solar PV data cannot be readily curated from existing public sources and if crowdsourcing alone may not scale up to enable frequent updates for a national-scale database of small-scale solar PV, another option is to use satellite and aerial imagery to perform this assessment. As we will discuss in Section II, numerous methodologies have been put forward for automatically identifying solar PV in satellite and aerial imagery using computer vision techniques. The benefit of these techniques is that with access to high-resolution imagery, these techniques can find and estimate the capacity of visible solar PV installations. In principle, PV installations should all be visible from above since this enables the sun to effectively hit their surfaces. Once properly trained, these methodologies can be applied to any imagery, and as updated imagery become available, the methods can be re-applied to rapidly obtain updated data. This narrative belies a key consideration: the reported performance of these techniques may not be representative of the algorithm's performance in general, but may be extremely specific to a particular scope of application. If so, we may not yet have a reasonable estimate of generalization performance of any previously reported solar PV detection algorithm (the authors' own past works included).

### *B. Challenges in accurately evaluating and comparing the performance of practical, automated PV assessment models*

Since the body of work on solar PV detection algorithms (as detailed in Section II) has shown that these methodologies can be effective in select pilot settings, the next logical step is to explore how well they will generalize broadly and scale up to practical applications. Solar PV identification algorithms typically rely on a training dataset from which the algorithm learns key patterns in the data and applies what was learned to make predictions on a test dataset; however, the experiment design may unintentionally bias the estimates of generalization performance (how it will perform on unseen data) or render the experiment incomparable with other methods.

#### *1) Potential source of biased performance estimates from test datasets: distribution shift*

One potential source of biased performance estimates is distribution shift caused by differences (e.g. differences in geography) in the statistics of the data used for performance evaluation and where these techniques will be applied in practice. For example, if we tested an algorithm on data from California and achieved outstanding performance, that provides no guarantees for performance in Germany, Shanghai, or any other location unless it is statistically similar to California. More subtly, the evaluation of the performance may depend on which subset of California was included. If the test region was the city of Fresno, CA, that may be sufficiently different from San Francisco, CA, that performance would not be representative. Even if the test data were from Fresno, if the test data happened to be sample that only contained dense urban areas within Fresno, the algorithm may not perform well on suburban Fresno imagery.

In addition to differences in geography, this same principle applies to any other systematic difference between the test data used for performance evaluation vs the data the algorithm will be applied to in practice. Such differences could arise from the choice of sensor modality (e.g. different satellites, satellites vs overhead imagery), temporal differences in the data (e.g. the test data are all from summer and the data in practice were all from winter), and even from differences in the atmospheric conditions at the time each set of images were taken. Each of these conditions may impact the statistics of the data and lead to biased performance evaluation.

*2) Potential source of biased performance estimates from test datasets: test data quality*

Another potential source of bias is relevant to the quality and trustworthiness of our test dataset. We typically assume that the test data, which we often refer to as "ground truth," are more or less infallible. However, for solar PV mapping, we rarely have a list of known locations of solar PV (otherwise, the analysis would be unnecessary). Therefore, we typically hire annotators to carefully scan overhead imagery for solar PV and draw outlines around the solar PV installations' boundaries. While these annotations are typically verified through the deployment of multiple annotators to ensure agreement, these annotators will have an error rate that is unknowable without known locations. Evaluating the quality of the ground truth ensure our test data are accurate for high quality performance evaluation metrics.

*3) Potential hindrances to performance estimate comparison across studies: level of spatial aggregation*

Heterogeneous levels of spatial aggregation (e.g. pixel-level, parcel-level, regional-level) of performance evaluation can at best, lead to results that are not directly comparable and can, at worst, obscure the correct interpretation of results at these different levels of spatial aggregation. Different levels of spatial aggregation indicate the resolution at which solar PV detection performance is evaluated. We typically divide performance metrics into two categories: *pixel-wise* and *object-wise* metrics. Pixel-wise metrics compare how each individual pixel in the image is classified (solar PV or not). In object-wise metrics, we consider each group of interconnected or closely-located pixels of solar PV to be a single object, since sometimes multiple separate solar PV panels that are near one another belong to one array (as on a building with a complex roof design). These object-wise metrics can then be aggregated further to parcel-level metrics (grouping solar PV located over one property as a single unit) or regional metrics (aggregating a neighborhood, city, county, etc. into a single number for evaluation). If our application requires accurately determining where each parcel-level installation is located, regional estimates may be insufficient. Similarly, if capacity estimates are required, measures that only evaluate the presence of solar PV but not their size (area) may be insufficient.

*4) Potential hindrances to performance estimate comparison across studies: proprietary data*

If the data are not publicly available, results from an analysis cannot be readily duplicated making it difficult to verify claims or build on past research. The use of novel and often unshared training and validation datasets, and the degree of diversity of the locations from which the training and validation data originate in terms of geography and sensor modality (e.g. different satellites, aerial photography) confound efforts to determine which methods are state-of-the-art. Depending on whether one problem is a bit harder/easier than another, this can lead to performance estimates that are not generalizable.

*C. Contributions of this work*

The purpose of this paper is to provide a comprehensive strategy for performance assessment of automated solar PV identification techniques for real-world applications, as well as to demonstrate the opportunities and challenges the process reveals, while challenging past practices empirically. Each of the above two sources of bias in estimates and the two hindrances to performance comparison impact how we understand the state-of-the-art in this field and these distortions of perspective may impact the efficacy of developing real-world applications and innovations relying on automated solar PV identification techniques. We explore these sources of bias and comparison hindrances and offer recommendations to remediate them. To that end, our specific contributions are as follows:

1. **Demonstrating performance across two domains (Section V)**. While most studies assume data are from similar geographic areas and use the same sensor (i.e. particular satellite), we explore the impact of applying a single training procedure when the training and testing data are different. We demonstrate that significant performance differences are possible and may be dramatic.

2. **Validating performance on ground truth from non-annotation ground truth and evaluating the quality of ground truth data (Section V.B-C)**. In past studies, evaluations were primarily based on manually annotated ground truth. We demonstrate the first automated solar PV performance evaluation that also includes actual data on known solar PV locations rather than solely evaluating performance based on human annotations of imagery data. We also compare the human-annotated satellite imagery data with the known locations of solar PV to assess the quality of the human-annotation process. Lastly, we evaluate the quality of the dataset of "known" solar PV locations and show that even that dataset has its imperfections.
3. **Investigating array-level performance evaluation and capacity estimation (Section V.D)**. We use array-level (nearby collections of solar PV on a single parcel of land) evaluation metrics to conduct a comprehensive evaluation of solar PV identification performance, including the size of the array. We present a new methodology for parcel-level grouping of detected solar PV arrays to achieve that goal. We also evaluate the potential of automated solar PV capacity estimation by building a regression model between aggregate estimates of municipal solar PV and total solar PV capacity of the municipality.
4. **Open code, open data, and an interactive map of prediction results for reproducibility and transparency**. We openly provide the SolarMapper framework for solar PV identification from overhead imagery. The framework involves pre-training on a large dataset (over 14,000 km$^2$) of overhead imagery and fine-tuning using a small amount of imagery from anywhere the framework needs to be deployed. We also publicly release the codebase [9], the input data [10] for training the algorithm in Connecticut, as well as results visualization for the entire state of Connecticut shown on the imagery used for prediction[1].

### *D. Organization of the paper*

We begin with a brief review of recent related research in Section II and present the technical details of the SolarMapper framework in Section III. We describe an array of complementary performance evaluation strategies that can be applied to the evaluation of automated solar PV array identification in satellite imagery in Section IV. Using those evaluation techniques, the key questions of performance evaluation outlined above are investigated in Section V; we where we analyze the impact of applying SolarMapper to new geographic locations and spatial scales. The experimental design investigates performance in the state of Connecticut and compares performance to San Diego, California (collectively an area covering more over 14,000 km$^2$), mapping individual solar arrays within those regions. We also explore performance at each residential parcel (i.e., each unique address) using proprietary ground truth data, and compare to manually annotated imagery to evaluate the comparative quality of ground truth. We summarize our findings, conclusions, and recommendations in Section VI.

## II. Related Work

The idea of using computer algorithms to automatically detect solar arrays in VHR imagery was first investigated in [11] (on a small-scale dataset) and [12] (on a larger scale dataset). These initial PV detection algorithms were designed using traditional image recognition approaches, consisting of hand-crafted image features and supervised classifiers [11,12]. These algorithms demonstrated the concept of mapping solar arrays in overhead imagery, but did not achieve performance that was likely to be useful in most applications.

Recently, convolutional neural networks (CNNs) have yielded groundbreaking recognition performance on many image recognition tasks [13], and CNNs were subsequently applied for solar array mapping [14–18]. CNNs were originally designed to provide a single prediction for an entire input image, e.g., indicating whether an image contains or does not contain a solar array. Of note was the work in [14,15,18] that employed semantic segmentation CNNs, which are designed to provide pixel-wise labels of an input image. In this work, and the context of remote sensing, we have referred to the semantic segmentation task as mapping. Substantially better performance was achieved for both solar array object identification, as well as estimating their shape/size, when using semantic segmentation models [15,17–24].

These studies using semantic segmentation (or mapping) CNNs demonstrated that solar mapping could achieve levels of performance useful in practice. Additional work around the same time demonstrated the possibility of inferring energy generation capacity using only overhead imagery [24,25]. This prior work collectively demonstrated the potential to create a system for reliably collecting small-scale solar PV information using remotely sensed imagery.

[1] Interactive visualization of identified solar PV across the state of Connecticut: https://energydatalab.github.io/solarMapper/

Some recent studies mapped solar PV installations using CNNs over large geographic regions, i.e. Hou et al. mapped 439 solar farms throughout China [21] and Kruitwagen et al. mapped over 68,000 commercial-, industrial-, and utility-scale PV facilities globally [18]. However, these studies were not focused on assessing small-scale rooftop residential solar PV, which accounts for 40% of the global solar PV installed capacity [26]. Most recently, Mayer et al. performed both classification and segmentation for residential solar PV identification on a real-world data covering 34,000 $km^2$ area in Germany [24].

As we discussed earlier, a key limitation of past work has been the ability to sufficiently evaluate performance to inform practical application of these techniques. The components described in detail above were (1) the use of manual annotations as ground truth rather than known PV locations, (2) the lack of evaluation of the impacts of distribution shift from changes in geography and imagery sensor modality through the use of optimistically homogeneous validation datasets, and (3) for large-scale studies, evaluating performance at an aggregate level rather than at the level of an individual solar PV array.

In past studies which mapped solar PV, the geographic scope was usually small [14–16,20,22,23], although some recent studies were performed on larger geographic areas [17,18,24]. These groups had limitations in the evaluation of performance which could impact the perception of the algorithm's potential for practical application. First, research that has investigated solar PV at the *individual array-level* [17,18,24] have relied on manual annotations of solar PV for ground truth. These annotations are based on expert judgement by human annotators – past studies have not used data on actual solar PV locations and there has been no quality assessment of solar PV array annotations that involves known solar PV installations. Second, distribution shift is a known challenge [27] in the computer vision community and is present whenever the validation data is in some way statistically different from the training data. This is common in solar PV assessment when the real world test imagery is from a significantly different location than the imagery used in training and validation or when the sensor used to collect the two were different (e.g. two different satellites or satellite imagery and aerial photography). However, in practice, the test data used for validation is often from the same distribution as the training data, so the detrimental impacts of distribution shift of not included in the evaluation of performance.

Past studies also overlooked some nuances of the problem of automated solar PV mapping by treating every object detected as a separate solar PV array, directly applying generic object detection and semantic segmentation evaluation metrics when evaluating the performance. This approach can lead to mistakes in the performance evaluation of installed solar PV. Consider the case when a building has multiple sets of solar panels upon it to accommodate the contours of the roof. Simply counting each array independently as a separate object [15] (rather than as a single building's solar PV installation) might result in inaccuracies in solar PV assessment. We seek to resolve this issue with array-level solar array groupings, which we present in this work.

## III. SolarMapper: A Framework For Mapping Solar Arrays In Overhead Imagery

### A. *Overview of SolarMapper*

We introduce our framework for solar PV identification in overhead imagery as a model built on successful ideas in the literature that we can use demonstrate performance evaluation challenges and evaluate strategies to overcome those challenges. Unlike most studies presenting a new solar PV mapping algorithm, we make no claims of performance superiority of SolarMapper; instead, this study is intended to clearly demonstrate the difficulty of accurate generalization performance evaluation and inter-study performance comparison and offer recommendations for overcoming these obstacles.

As described previously, our goals are to (a) investigate the impact of domain shift on the performance of similarly trained methods, (b) evaluate the quality of ground truth and its impact on performance, (c) investigate the results of different types and spatial resolutions of validation on the interpretation of the quality of methods performance, and (d) share content to maximize transparency of performance and ease of reproduction. Towards that end, we need a consistent testbed for implementing our experiments for (a), (b), and (c), as illustrated in Figure 1. We use one consistent experimental design, as we describe in the paragraphs that follow to ensure consistency of implementation.

The core of SolarMapper is a CNN that has been trained to recognize solar arrays in overhead imagery. SolarMapper receives an overhead image as input, and returns a confidence "map", indicating the likelihood that a solar array exists at each pixel location in the original image. To obtain a categorical label at each pixel – "panel" or "not a panel" - we can apply a threshold to each pixel value (e.g., 0.5), above which a pixel is assigned a label of one (panel), and otherwise zero (non-panel).

To make SolarMapper capable of mapping solar PV at any given location and geography, we adopt a two-step approach. The first step is the pre-training where we train SolarMapper on a large manually annotated dataset so it develops some basic capability of making pixel-wise solar PV prediction. The mapping capability of the pre-trained model is optimized on its training imagery, but in reality, may produce poor results when applied to imagery whose characteristics are likely to be different from those of the training location. To address this issue, we perform the second step where we use local imagery from the location of interest to fine-tune the network. In contrast to the pre-training, the fine-tuning usually requires much less data since the network doesn't learn to map solar PV from scratch but only needs to adapt to new imagery characteristics it hasn't seen before. We describe more details about our two-step approach in the following sub-sections and in appendices. We also provide an overview of each step of the SolarMapper experimental design in Figure 1.

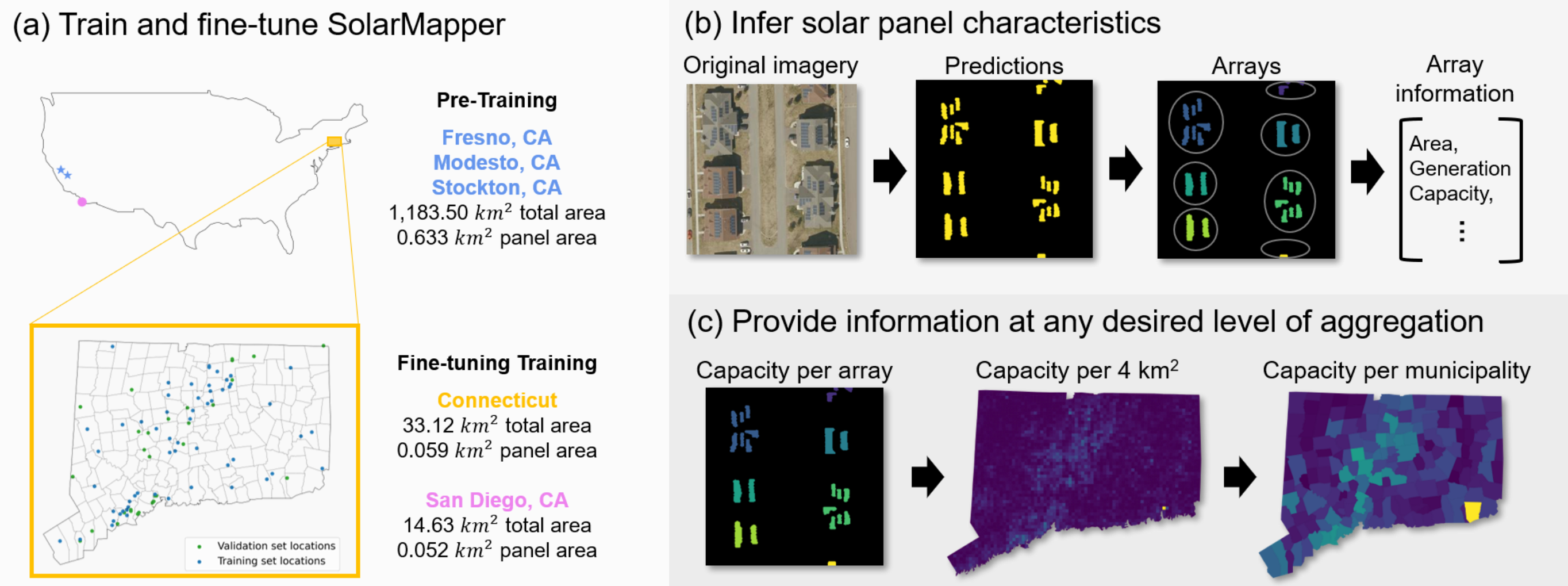

**Figure 1. An illustration of the major steps of our proposed approach for large-scale mapping of solar array information using the proposed SolarMapper mapping tool. The software for SolarMapper is released publicly, along with procedures for how to apply it to new locations.**

## *B. Pre-training SolarMapper*

Training a CNN requires a set of imagery, termed a training dataset, for which the true labels of each pixel are known to the network. CNNs consist of (often) millions of parameters that each influence their outputs; training a CNN involves iteratively adjusting these parameters until it produces accurate labels for the training dataset. To achieve the best performance, CNNs require training datasets that are large and representative of the desired application domain (the location to which you would want to apply the CNN), and its diversity. SolarMapper was trained on the Duke California Solar Array dataset [17], comprising over 1,000 $km^2$ of imagery across three cities in the US state of California, and encompassing 16,574 hand-labeled solar arrays (see Appendix I for dataset details). To date, this dataset is the largest and most diverse publicly-available dataset of fully-annotated solar arrays.

We employed a two-fold cross-validation procedure on the Duke California Solar Array Dataset to search for the best network training parameters including network architecture, learning rate, etc. Cross-validation is a conventional procedure within the machine learning community to estimate the performance of supervised (i.e., trained) models, such as CNNs. Having found the best training parameters, we then trained a new model using the entire Duke California Solar Array Dataset to maximize the usage of labeled training data. Full technical details about the training of SolarMapper can be found in Appendix II.

## *C. Local fine-tuning of SolarMapper*

Applying SolarMapper to a new location is not a trivial endeavor. This is because of the high likelihood that differences may exist in the characteristics of the imagery at a new location compared with the imagery of the training dataset. Such qualitative differences can cause SolarMapper to perform (i) unpredictably and (ii) poorly, making it unusable for most practical applications.

Such changes in the imagery may arise due to changes in: underlying landscapes (e.g., vegetation vs desert), appearance of urban structures on which solar arrays reside (e.g., roof colors, textures), lighting conditions (e.g., due to changes in hour of day), sensor modality (e.g. different satellites, aerial imagery, drone imagery), and more. This is a problem shared by all supervised machine learning algorithms and a major ongoing challenge recognized for remote sensing applications in particular [29,30].

Fine-tuning is a potential solution to this problem. In our context, fine-tuning aims to leverage the similarity between the tasks of finding solar arrays in CA (the training data for SolarMapper) and finding solar arrays in CT (our target task, and likewise for San Diego, our second target task). The idea is that SolarMapper's parameters after pre-training require minor adjustments to perform well in CT. With a relatively small amount of local hand-labeled data we may adapt, or fine-tune, the parameters, thereby achieving highly accurate results with only a small fraction of the training data required to train a full CNN. The process of fine-tuning for CT is illustrated in Figure 1. For our fine-tuning experiments, we hand-annotated about 50.56 $km^2$ of imagery in CT and 22.50 $km^2$ of imagery in San Diego, corresponding to 0.4% and 33% of the total available imagery at each location respectively, but representing approximately the same fixed-budget of human annotation time in each case. To evaluate SolarMapper's performance after fine-tuning at each location, we employed a 2:1 training to validation split: we used about 2/3 of labeled imagery to fine-tune SolarMapper and validated on the remaining 1/3 of the imagery. SolarMapper was fine-tuned at each location separately, so two fine-tuned models were obtained and their respective performance evaluations are in Table 3. Full technical details can be found in Appendix B.D.

The idea behind this approach is to create a structured, practical application of an algorithm pre-trained on one large dataset and fine-tuned for practical application in two locations, keeping the fine-tuning process the same. Through the evaluation of performance for these locations, we discuss what we learn about how well these techniques work in different regions (i.e., performance generalization) and evaluating the difficulties of performance comparison to other techniques.

## IV. Evaluation Metrics Of Solar PV Mapping

At the core of this work is the performance evaluation. Existing studies usually involve performance evaluation at one of two levels of aggregation: pixel-wise segmentation and object-wise detection. Pixel-wise segmentation evaluation is performed directly on the pixel-wise solar PV segmentation results and evaluates the quality of each pixel-wise prediction of the shape and size of the solar PV present in the image. In contrast, object-wise detection evaluation is performed on objects formed by groups of individual pixels and it measures how well a model predicts the occurrences of solar PV as objects regardless of the size or the shape predicted.

These two approaches are actually quite different and each appropriate for different applications. If we care about the total installed capacity of solar PV, then pixel-wise metrics that measure how well we classified all the pixels in each solar array would be more appropriate (like intersection-over-union or IoU). If we care about counts of solar arrays, then object-based metrics are appropriate (like object-wise precision and recall). Additionally, for these object-wise metrics, if we care about estimating the capacity (in kWh) of each *rooftop installation*, then we need not just standard object-wise metrics, which is typical of the papers previously discussed in this space, but we need building-wise or parcel-wise metrics, treating each close grouping of solar arrays as a single installation.

Since different applications require different metrics, providing a collection of metrics for a given problem will be more useful to a wider swath of practitioners and researchers than providing one type of metrics, alone. In this section, we explore these metrics in this section and present a new approach to grouping collections of solar arrays into installations / parcel-level collections.

### A. Pixel-wise segmentation evaluation metrics

Pixel-wise predictions are the direct output of any segmentation based solar PV mapping method. To evaluate pixel-wise performance we use the intersection-over-union (IoU) metric, which is popular for scoring pixel-wise labeling tasks (often called semantic segmentation) in the computer vision research community [31,32]. Given two sets of pixels denoted by A (e.g., predicted panel pixels) and B (e.g., true panel pixels), IoU is given by $\mathrm{IOU} = \frac{|A \cap B|}{|A \cup B|}$. Here the vertical bars indicate the cardinality of the set (for our case, the number of pixels in each set of pixels). An IoU of one is achieved if the predicted array pixels perfectly overlap with the true array pixels. If there is no overlap, the IoU will be zero. This metric is particularly useful in evaluating how accurate the estimates of power capacity associated with each solar array will be since the capacity of an array is proportional to the area of that array.

### B. Array-wise detection evaluation

In many practical applications, the information on the occurrences and locations of solar PV is desired as much as the information on sizes and shapes. In the computer vision research community, object-wise detection describes the task of finding occurrences and locations of the subject of interest. However, segmentation-based solar PV mapping does not produce object-wise predictions directly. So, object-wise predictions can only be obtained from some grouping procedure based on pixel-wise predictions. When making object-wise predictions of solar PV from pixel-wise predictions, past studies usually label each group of connected pixels an object and consider each labeled object a solar PV *panel*. However, it is common that multiple panels are installed on the rooftop of one structure (as shown in Figure 2(a));we term all panels located on the same structure an *array*.

We argue that grouping objects for detection performance evaluation by *arrays* rather than *panels* is better for solar PV mapping because:

1. *Array*-wise grouping adds another dimension for residential solar PV related analyses and could be more compatible with other types of data like household-level census data and building area data.
2. *Panel*-wise grouping is less robust when pixel predictions for a panel are not perfectly connected to be grouped as one predicted panel.

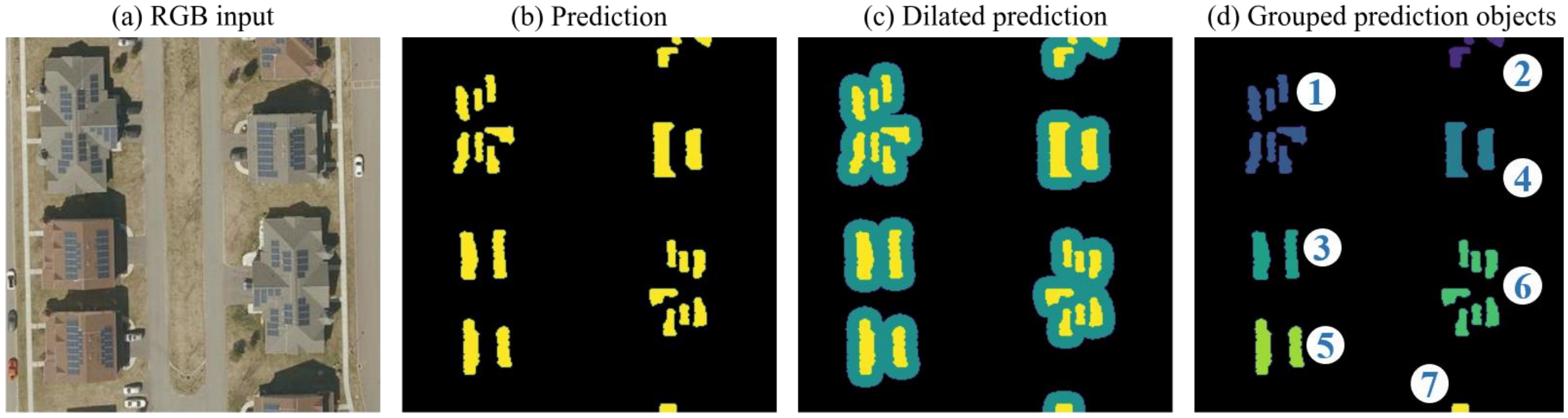

**Figure 2. Dilation-based array grouping process. Arrays in (a) all consists of multiple panels. We apply a 3-meter dilation to create (c) dilated prediction from (b) prediction panel pixels. Grouping connected components in (c), 7 multi-panel arrays can be identified as labeled in (d).**

Having obtained predicted and ground truth arrays from the array-wise grouping procedure, we can then compare them with ground truth to evaluate the detection performance. We say that a predicted panel array is a true positive if it achieves an IoU $\geq 0.5$ with a true panel array. Otherwise, it is considered a false positive. This is a common criterion for detection within the computer vision community [33,34]. Any ground truth arrays that are not linked to a predicted array are considered false negatives. Based on this procedure, we can obtain the precision (proportion of detections that were true positives) and recall (proportion of true PV arrays that were detected). Precision and recall are common measures for object detectors [33,34], including in remote sensing applications [12,35]. By varying the confidence score threshold, we can obtain multiple pairs of precision and recall and plot a precision-recall (PR) curve. An average precision score (area under the PR curve) can then be calculated from a PR curve as a summary statistic, where the higher the average precision the better (up to a score of 1).

The detection of solar PV arrays or any other rare objects differs from most balanced classification tasks because the total number of ground truth objects could be relatively low. In this case, calculating precision does not always reflect how a model is prone to making false positives since it is tested on a limited number of solar-PV-looking objects. Therefore, we also report an improved metric, false positives per unit area, to adjust for this issue. Similar to plotting a PR curve, we vary confidence thresholds to obtain a curve as well. This curve is a modified receiver operating characteristics (ROC) curve, where true positive rate (recall) is on the vertical axis and the false positives rate *per unit area* is on the horizontal axis.

### C. Evaluation by residential and commercial solar PV arrays

While we focus on the smallest common category of solar PV, distributed residential solar PV, some commercial solar PV arrays are included in the imagery, inevitably. We notice that commercial PV usually appears quite different in visual characteristics than residential PV. Commercial PV is usually larger and sometimes located in less populated area, both of which could affect the difficulty of the mapping task significantly. Therefore, we evaluate residential and commercial solar PV mapping performance separately to ensure that reported metrics are as comparable as possible.

We use a simple area-based approach to separate residential and commercial PV arrays where we (1) first perform the array-wise grouping procedure on the pixel-wise ground truth to obtain ground truth arrays; and (2) classify residential and commercial arrays with a capacity threshold of 20 kW [36] (equivalent to $100\ m^2$, see Appendix C. for details) on grouped arrays. Ground truth arrays whose total sizes are smaller than the size threshold are designated as residential PV arrays and the rest are commercial PV arrays.

### *D. Evaluation by housing density in surrounding areas*

Another potential different from one region to another is that population density may vary from region to region and this may impact the number of examples present. Some solar PV arrays are located in rural areas while some are located in more densely populated areas with all different types of buildings blended together. When mapping solar PV over large geographic areas, such difference in surrounding environments could impact the comparative performance evaluation metrics. To account for this discrepancy, we perform stratified evaluation where we put ground truth arrays into 3 brackets by the regional housing density for each image used in this study. The housing density information is inferred from Microsoft's US Building Footprints [37] dataset and more technical details could be found in Appendix D.

## V. PERFORMANCE EVALUATION: RESULTS AND DISCUSSION

In this section, we dive deeply into each of the performance evaluation strategies previously discussed and their implications for generalizability.

### *A. Validation on manually annotated ground truth: distribution shift on display*

Qualitatively, we present the results of this analysis in Figure 3, showing examples of residential, commercial, and industrial solar PV arrays across (a) Connecticut and (b) San Diego, CA and the varying quality of annotations in these locations. There are two takeaways from these examples. First, while these data are nominally of the same resolution, the CT data is clearer. Secondly, we can see that generally, the CT predictions were more accurate (had more true positive pixels) than those in San Diego. However, these were pretrained using the same data and fine-tuned for the region (CT or San Diego, respectively) using the same procedure.

Quantitatively, summary statistics at both locations showed that the model had dramatically better performance pixel-wise and array-wise in CT than in San Diego (Table 1). The imagery resolution was the same across these regions, however, but the sensor and geography were different. These differences are common across different studies in the field as well. Even though IoU is a common metric in evaluating pixel-wise semantic segmentation performance, it is therefore hard to compare one study's IoU metrics directly with another study's since similar differences in the imagery or geography are typically present. One recent study, where the imagery resolution is the same as in our study (0.3 m/pixel), reported IoUs ranging from 0.5980 to 0.6249 when fine-tuning a pre-trained model to new locations [22].

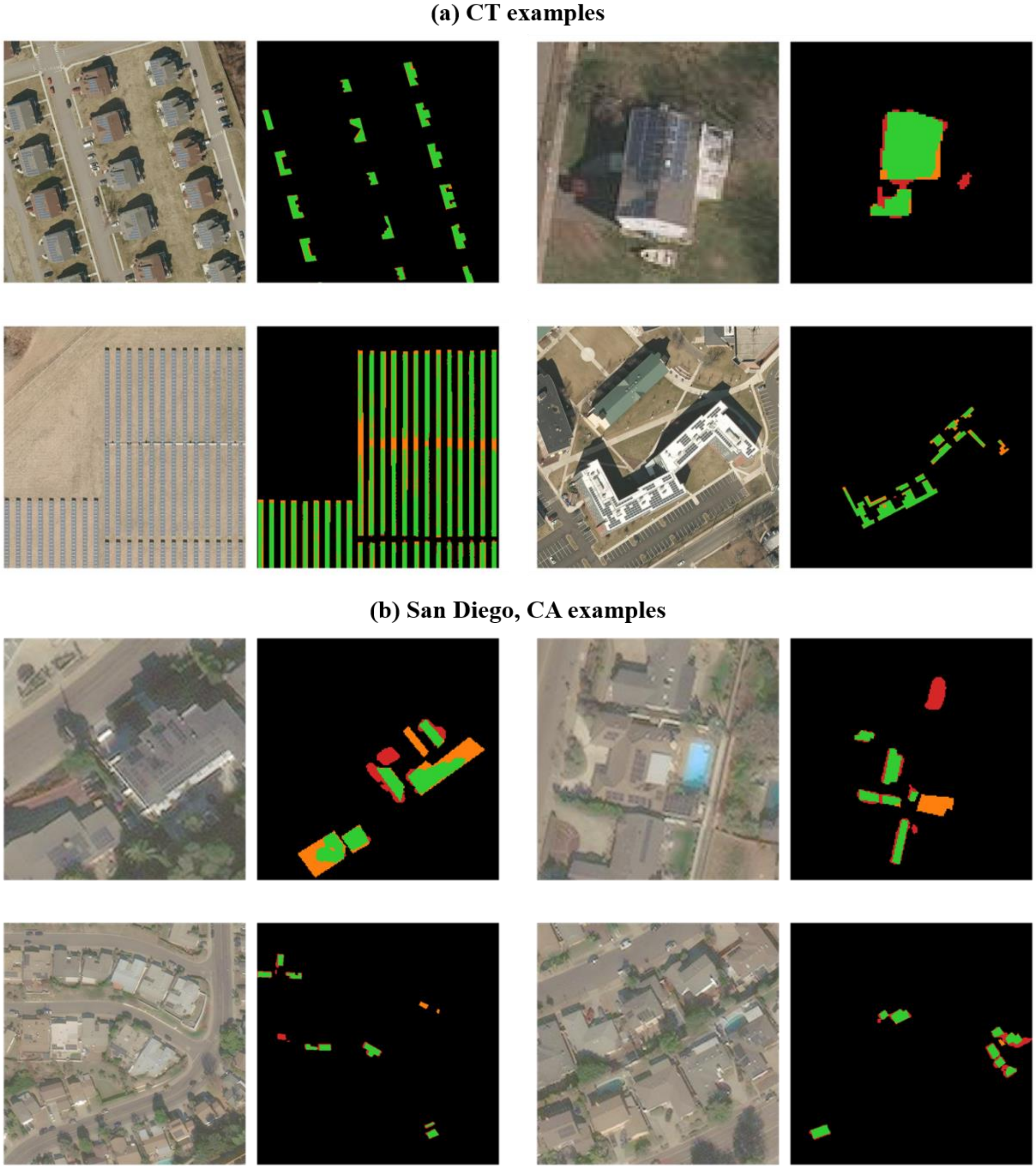

**Green**: true positives. **Red**: false positives. **Orange**: false negatives.

**Figure 3. Pixel-wise prediction examples.**

**Table 1: Pixel-wise and array-wise fine-tuning performance of SolarMapper on CT and San Diego, CA validation dataset.**

| Fine-tuning Location | Pixel-wise IoU | Array-wise Overall AP | Array-wise Residential AP | Array-wise Commercial AP |
|---|---|---|---|---|
| CT | 0.7326 | 0.8236 | 0.8179 | 0.8912 |
| San Diego, SD | 0.6005 | 0.7089 | - | - |

Additionally, in Figure 4, the accuracy of the annotation is measured with IoU – generally, the higher the IoU, the more likely a correct prediction of power capacity can be made because the area of the PV array will be accurate. Figure 4(a) shows the distribution of IoU for all, residential, and commercial arrays, and most arrays were predicted with an IoU of 0.8-0.9, which is considered quite high in the computer vision literature (E.g. some recent solar PV mapping paper report their best IoU to be around 0.6-0.7 [22]). Figure 4(b) shows that larger PV arrays based on size tended to have larger IoU values. In contrast, Figure 4(c) shows the distribution of IoU for residential arrays in SD (no commercial arrays found in SD validation imagery), and most arrays were predicted with an IoU of 0.6-0.7, which is lower than what Figure 4(a) reveals in CT. Figure 4(d) also shows more arrays have IoUs around 0.6-0.7 compared to the case in CT. These results further illustrate the problem of distribution shifts as different building types (residential vs commercial) and different geographies (CT vs SD) result in different levels of performance.

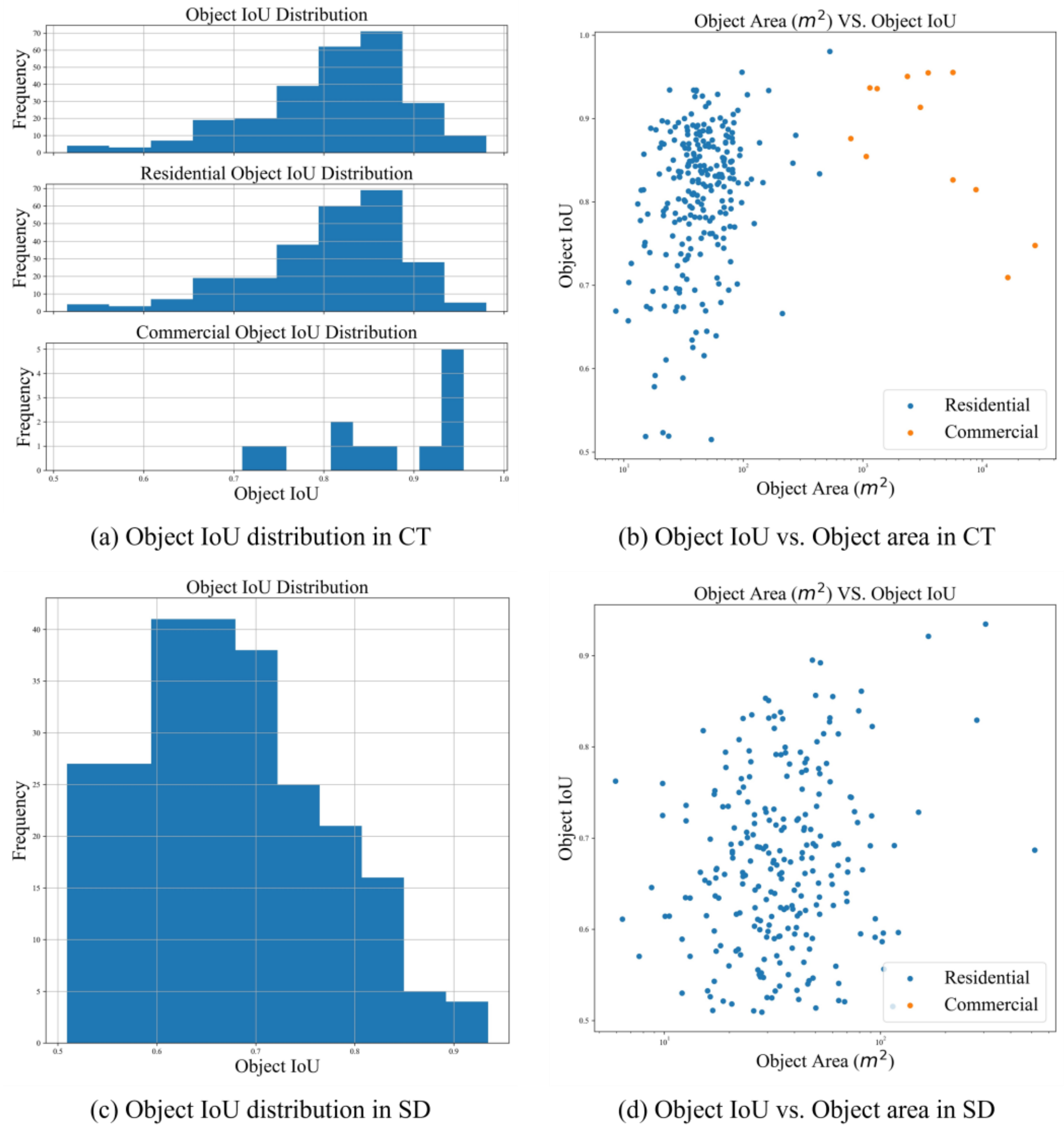

(a) Object IoU distribution in CT

(b) Object IoU vs. Object area in CT

(c) Object IoU distribution in SD

(d) Object IoU vs. Object area in SD

**Figure 4. Array-wise IoU analysis for Connecticut validation set.**

Array-wise PR and ROC curves for the validation images at both locations are shown in Figure 5. The model at CT generally shows vastly superior performance compared to SD in terms of higher average precision (AP) (see Figure 5(a)), higher true positive rate (recall) and lower false positive per unit area (see Figure 5(b)). Even the worst-case scenario in CT provides a true positive rate around 0.9 at a low false positive rate of 5 per $km^2$. Most false positives can be rapidly filtered by manual inspection after prediction. This approach captures over 90% of the objects of interest and the false positive rate is manageable for screening post-processing. Comparing sub-groups of CT validation images, the performance generally improves as the housing density decreases, indicating that detecting solar PV panels at more densely populated area is intrinsically a more difficult task. However, CT's high housing density performance stills shows a non-trivial advantage compared to SD performance in terms of higher maximum recall and much lower false positive rate. This may suggest that although housing density in surrounding areas has a big impact, additional factors contribute to the model's superior performance in CT compared to SD, which could be a major challenge with some practical use of the framework or any other mapping tool in overhead imagery.

***Key finding***: These results clearly demonstrate that with equivalent model training and highly symmetric fine-tuning processes, vastly different performance results are possible when the test data are different. These models are not inherently robust across geographies or urban density differences.

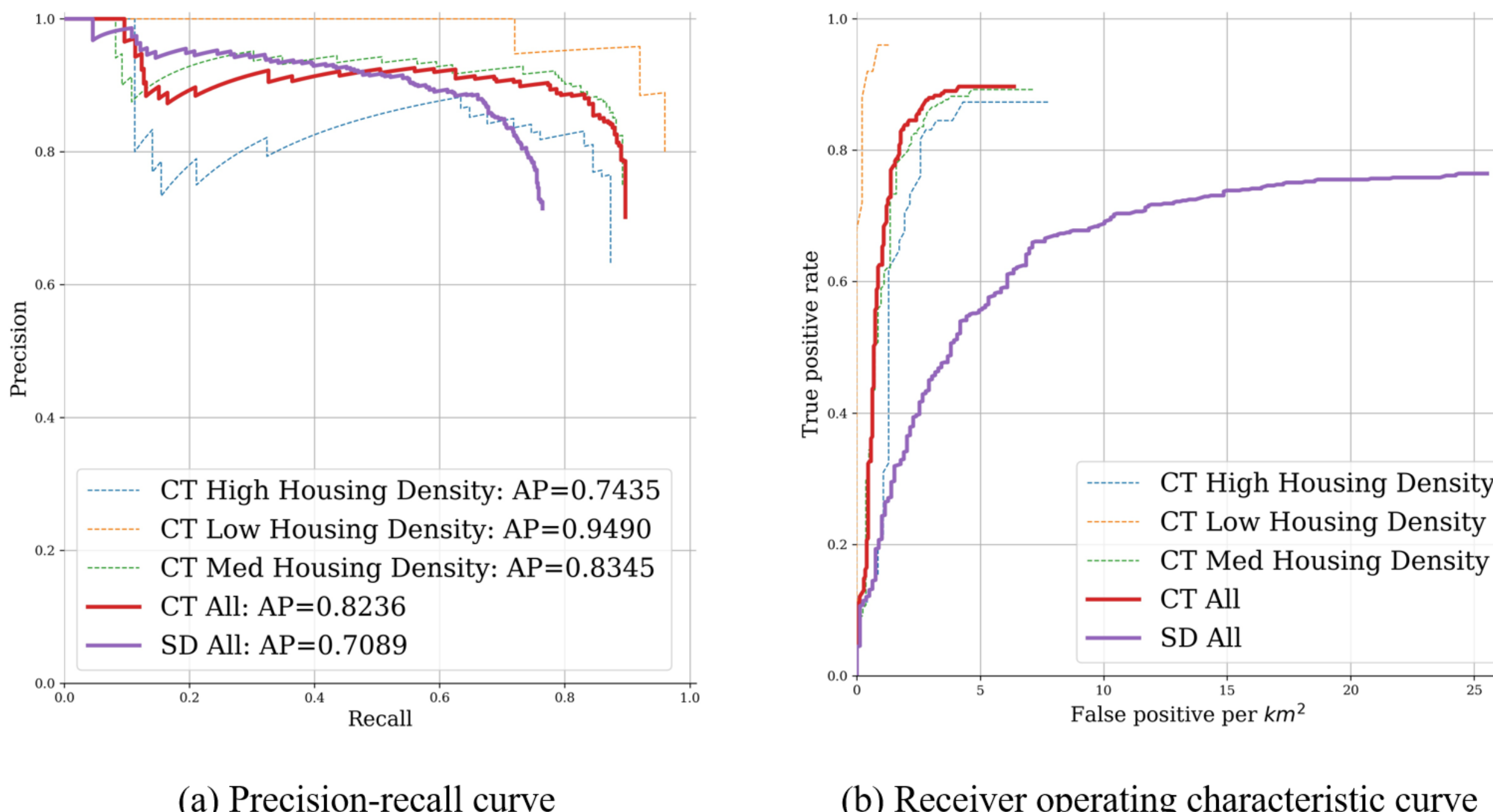

(a) Precision-recall curve

(b) Receiver operating characteristic curve

**Figure 5. Array-wise evaluation for CT and SD.**

### *B. Validation on ground truth from solar installer data*

In this section, we evaluate SolarMapper's effectiveness in identifying solar panels by validating its results with records from a parcel level residential solar panel array database in Connecticut. Additional data sources used for this validation include residential parcel boundaries from Digital Map Products (DMP) [38] and parcel-level solar panel array data for the entire state of Connecticut from Lawrence Berkeley National Laboratory's Tracking the Sun (TTS) [36] series. In this section, we also evaluate the quality of this parcel level data. Since it is rare to have access to such information, we investigate the quality of the data as it compares to data from manually annotated imagery.

#### *1) Parcel-level detection prediction performance*

We begin by comparing our predictions from SolarMapper to the ground truth at the parcel level. To do that, we needed first to match the TTS solar arrays with the DMP parcels. We focus the analysis on areas where both the DMP and the TTS datasets have coverage. We first compared the DMP and the TTS datasets spatially and identified DMP residential parcels that contained any TTS points: we called these TTS parcels. Then, we spatially joined the TTS parcels with our predicted solar PV arrays. We performed a proximity matching where we called a prediction as a true positive (TP) when the centroid of the predicted solar PV array polygon(s) fell within 5 meters from the TTS point in a TTS parcel. If the centroid of the predicted array polygon(s) fell more than 5 meters from any TTS point, we called it a false positive (FP). If no predicted polygons matched with a TTS point by proximity matching, we called this TTS point a false negative (FN).

**Table 2. Results summary of parcel-level matching analysis in Connecticut.**

| | TTS points | Predicted arrays |
|---|---|---|
| Total | 15,838 | 25,501 |
| Matched | 12,115 | 11,570 |
| Unmatched | 3,723 | 13,931 |
| Precision | 0.4537 | |
| Recall | 0.7649 | |

Using these values, we can calculate a raw score of precision and recall based on the matched points compared to the TTS point, which leads to a precision of 0.45 and recall of 0.76. However, before we put too much trust in those performance estimates, we need to first verify the TTS data, approaching each of our "ground truth" estimates with caution.

*2) Challenges with the solar PV array reported data*

Approaching the ground truth data with caution, we can use manual inspection of the imagery data, which is contemporaneous with the TTS data. We manually inspected 400 instances for each of the three types of matching results (true positives, false positives, and false negatives) to verify the quality of the TTS data. For all 400 true positives, there was a clearly visible solar array in each. However, we found that 42% of the false positives and 67% of the false negatives were *actually correct predictions* made by the model, demonstrating that the TTS points dataset itself included inaccuracies. These inaccuracies can obviously impact the metrics calculated from raw matching of results presented above, effectively *underestimating* the performance of the model.

It is equally important to look for potential inaccuracies in true negatives. True negatives are ill-defined in the case of object-wise performance evaluation since everything that is not a TP, FP, or FN is essentially a true negative. Therefore, to investigate the prevalence of true negatives, we manually inspected 4 image tiles covering about 2 $km^2$ of area and no visible errors in true negative assignment were identified.

Based on our manual evaluation of the ground truth, if we assume that the rates of error identified for TP, FP, FN, and TN were correct for Connecticut, then we can adjust precision and recall calculation for the parcel-level matching analysis to get a more accurate estimate of precision and recall.

*3) Precision and recall estimate adjustments to account for systematic errors in ground truth*

Through the manual inspection above, we estimated the error rate of false positives being erroneous ($ER_{FP}$) to be 0.42 and the error rate of false negatives being erroneous ($ER_{FN}$) to be 0.67. We assume these error rates are consistent throughout the entire state of CT since our sampling was randomized, and using that we estimated the total numbers of erroneous false positives ($E_{FP}$) and erroneous false negatives ($E_{FN}$) using Eqn (1). We had no need to update TP rates since we found no errors through inspection, but the calculation would be similar.

$$Err_{FP} = FP \times ER_{FP}, \qquad Err_{FN} = FN \times ER_{FN} \tag{1}$$

By definition, if a prediction is mistaken as a false positive, it should be corrected as a true positive. Therefore, the adjusted false positives ($FP_{adj}$) is the original false positives ($FP$) subtracted by the estimated erroneous false positives ($E_{FP}$). The subtracted erroneous false positives ($E_{FP}$) should then be added to the original true positives ($TP$) to get the adjusted true positives ($TP_{adj}$). Similarly, we can calculate the adjusted false negatives ($FN_{adj}$) by subtracting the erroneous false negatives ($Err_{FN}$) from the original false negatives ($FN$). These adjustments are summarized in Eqn (2).

$$TP_{adj} = TP + E_{FP}, \qquad FP_{adj} = FP - E_{FP}, \qquad FN_{adj} = FN - E_{FN} \tag{2}$$

Finally, we can get the adjusted precision and recall by applying Eqn (2) into Eqn (3).

$$Precision_{adj} = \frac{TP_{adj}}{TP_{adj} + FP_{adj}}, \qquad Recall_{adj} = \frac{TP_{adj}}{TP_{adj} + FN_{adj}} \tag{3}$$

$$Precision_{adj} = \frac{TP + E_{FP}}{TP + FP} \tag{4}$$

$$Recall_{adj} = \frac{TP + E_{FP}}{TP + E_{FP} + FN - E_{FN}} \tag{5}$$

Given the adjustment of metrics described above, the adjusted precision and recall for the parcel-level matching are 0.6689 and 0.9360, respectively, demonstrating a significant difference compared to our estimates from the raw data alone. These estimates are vastly improved over the raw estimates before the corrections are taken into account.

***Key finding***. The "gold standard" for solar ground truth is generally considered to be records of installations. We show that these data may also be incomplete and in ways that significantly impact perceptions around performance and therefore must be used carefully and cautious skepticism applied to their validity.

## *C. Manually annotated ground truth analysis*

In the same way that we applied a cautious skepticism to the TTS data, using the manual annotations to evaluate the TTS data, we perform the reverse analysis and use the TTS data to evaluate the manually annotated ground truth data. As a reminder, these hand-labeled image annotations are the same data that were used for fine-tuning and validation of performance in Connecticut. While manual labeling of ground truth is common in computer vision applications, it's possible that some challenging-to-identify objects could be sometimes missed and moreover, they could lead to degradation in performance [39]. Similarly, human annotators can also mistakenly label objects as solar arrays even if they are not; sometimes it is difficult to discern whether a particular object is truly a solar array or not (e.g., rooftop skylights).

We performed an evaluation of the manually-annotated ground truth following the same process as in the last section. For the 87 tiles we annotated in CT (see Table 3), we found 60 panels in total which are visible but didn't match with any predicted solar PV panel. This represents only about 3.8% of the panels we labeled in the selected region. Figure 6 shows some examples of missed panels.

**Table 3. TTS matching results with labeled ground truth**

| Total TTS points | | 294 |
| --- | --- | --- |
| TTS points matched with predictions | | 214 |
| Unmatched | Panels visible | 60 |
| | No panels visible | 20 |
| Total predicted solar PV panels | | 1,598 |

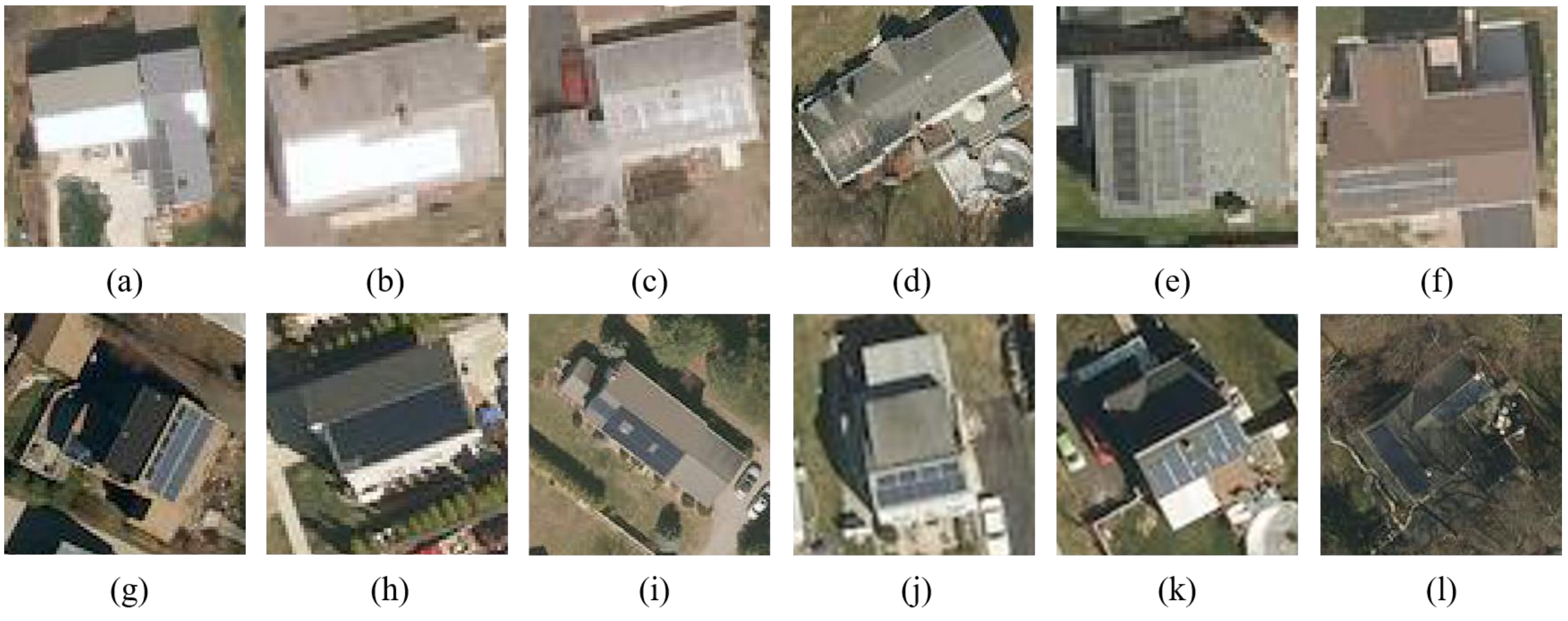

**Figure 6. Examples of solar panels missed by human annotators.**

***Key finding***. While the error in manual annotation was nonzero, it was considerably smaller than the similarly evaluated error in the TTS locations. Manual annotation using trained annotators may be highly effective for generating accurate ground truth, although it is unlikely to be 100% effective especially for hard-to-identify solar arrays.

## *D. Estimating performance (via power generation capacity estimates) in Connecticut at the municipality level*

So far, we have explored pixel-wise and array- / parcel-level performance of SolarMapper; in this section we demonstrate how SolarMapper can be evaluated at the municipality level and used to estimate power generation capacity over larger areas, using only the original overhead imagery, and the predictions provided by SolarMapper (derived from overhead imagery). We demonstrate the proposed approach in the state of CT, building on the results we have presented previously. We are uniquely able to validate our capacity predictions in CT through the data compiled by Data Driven Yale from the U.S. Census Bureau and individual municipalities. These estimates of the installed solar capacity in each of its 168 cities (termed municipalities), are provided via the Solar Scorecard Project [40]. Of course, given the discussion in the past two sections, we need to express caution around trusting any one dataset without verifying its quality. We account for issues with the data by using the relative estimates by municipality rather than the absolute estimates of capacity.

We begin by providing a brief overview of our general approach for capacity estimation, and then presenting the results.

*1) Estimating capacity from overhead imagery*

Our approach for inferring capacity relies on the relationship that the power generation capacity of a solar array, denoted $c$, is proportional to its surface area, denoted $a$. If we assume some uncertainty in the relationship, then we obtain the following simple linear regression model to predict the capacity of the $i^{th}$ array:

$$c_i = \gamma a_i + \eta_i. \tag{6}$$

Here $\gamma$ serves as a proportionality constant, indicating the capacity per unit of surface area. The value of $\gamma$ will likely vary for each solar array depending upon factors such as its manufacturer, its age and maintenance, and its composition type (e.g., thin film, polycrystalline, etc.); but it is approximated here as a constant across arrays.

Using PV array predictions from SolarMapper, we can estimate the (approximate) surface area of each array by summing the number of pixels. We have a known spatial extent of each pixel that can be used to estimate the area in square meters. This basic model and approach were demonstrated (with a different mapping algorithm) to yield accurate estimates of capacity for individual arrays in [25].

To employ this model in practice however, one must obtain an estimate for $\gamma$, and we propose two approaches. The first is to use prior information, perhaps from solar PV manufacturers, to estimate a likely value for $\gamma$. Alternatively, as in the method used in So et al. [25], we use a small set of known values of solar array capacity and area, estimated using SolarMapper, to infer $\gamma$. This can conceivably be accomplished using linear regression (as in So et al. [25]) with a very small number of $(c_i, a_i)$ samples.

In this work, we employ a modified version of the latter approach, in which we use municipal-level aggregated capacity values and surface areas to infer the regression coefficient $\gamma$.

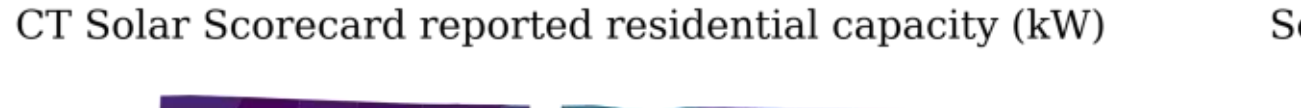

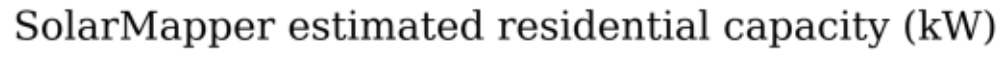

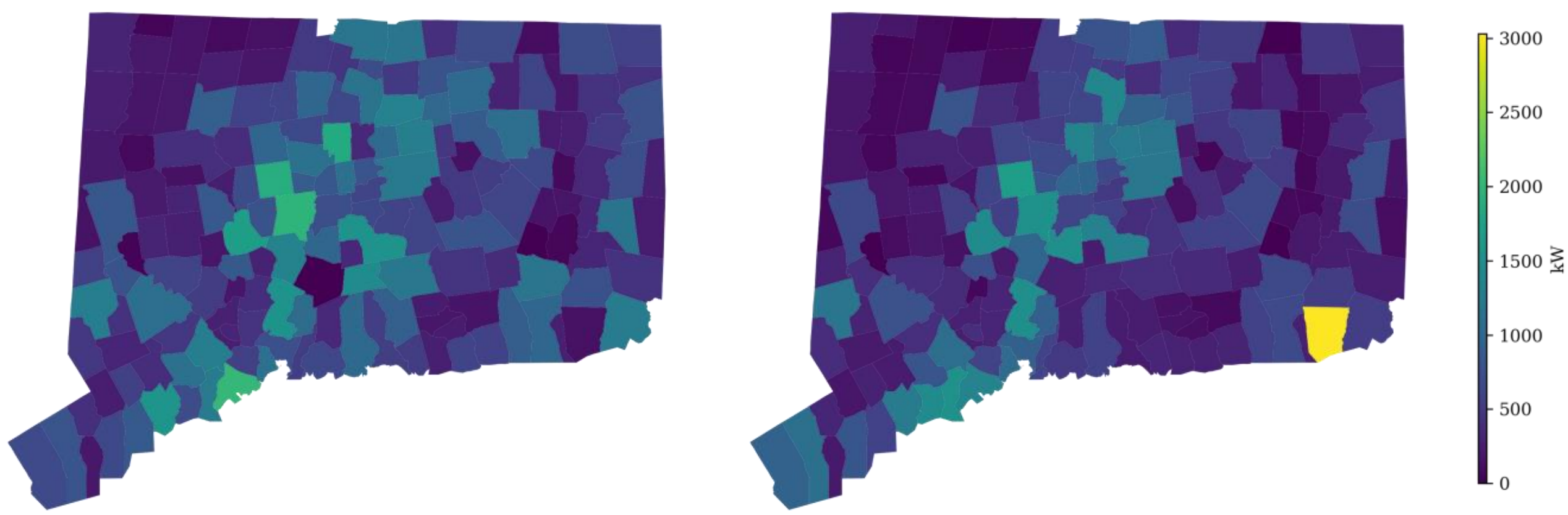

**Figure 7. Residential solar PV capacity: officially reported values (left) and SolarMapper estimated values (right)**

*2) Validating the capacity estimates*

After estimating the power capacity of each detected solar array in CT, we summed the capacity of all arrays within each municipal region. To evaluate the accuracy of our capacity estimates, we computed the Pearson correlation coefficient between our capacity estimates and those reported in the CT Solar Scorecard dataset, as shown in Figure 8.

If we use a model that assumes a fixed value of $\gamma$ for all arrays, we achieve a correlation coefficient of 0.7622. Using color imagery to estimate a unique value of $\gamma$ for each solar array results in a slightly higher correlation coefficient of 0.8859. In both cases the p-values were less than 0.01. Figure 7 presents a visualization of the estimates provided by SolarMapper, and the officially reported values. The results are visually consistent with the high correlation coefficients. We note in Figure 7 that one municipal region, Groton, was removed from our analysis because it is an outlier with a known cause due to an anomaly in the data. This is discussed in more detail in Appendix E.B.

The values of $\alpha_i$ used in equation (3) to estimate capacity are based directly on the number of pixels detected by SolarMapper. Therefore, it is unlikely that capacity estimation would be accurate unless SolarMapper provided accurate values of $\alpha_i$ for each panel. This analysis provides a practical benchmark of SolarMapper's performance, since it suggests that SolarMapper was sufficiently accurate to establish a high correlation with a known municipal capacity estimation dataset.

***Key finding***. Since capacity data are rarely available at the individual household or parcel level, evaluating the performance of capacity estimation will likely need to be completed in aggregate. Since the ground truth is likely to have flaws, checking for correlation between the estimates allows us to measure how aligned our estimates are with the (likely imperfect) ground truth. However, errors in even one data point in the "ground truth" can dramatically impact the reported performance.

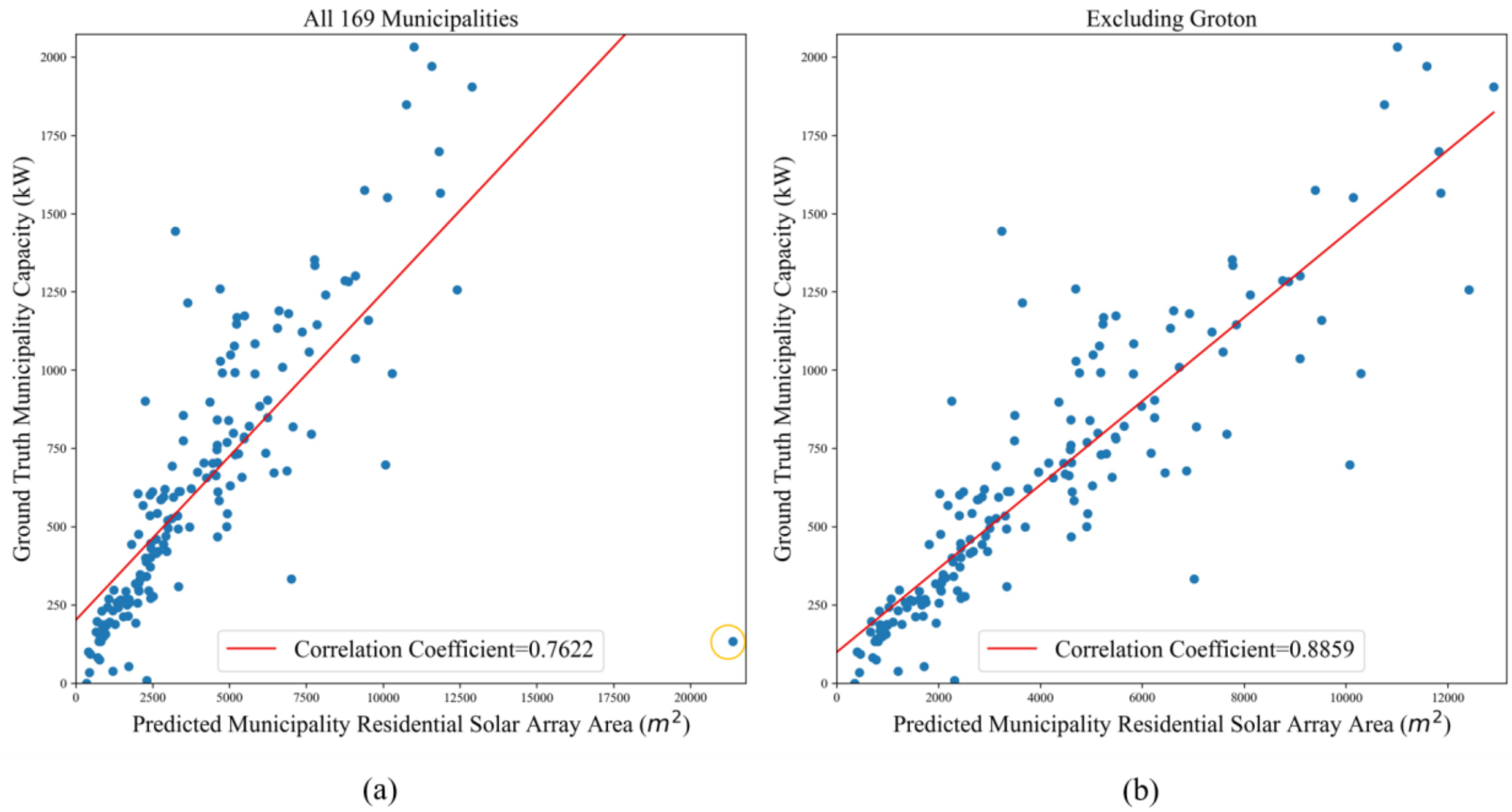

**Figure 8. (a) Installed small-scale solar PV capacity in CT and the residential solar array sizes estimated by SolarMapper for all 169 municipalities in CT. Upon investigation, Groton (highlighted in orange in (a) at bottom right corner) was found to have accurate array predictions, and was identified as an outlier (see Appendix IV). Groton was then removed from our experiments and see results in (b).**

# VI. CONCLUSIONS

While this is often the section where we would try to present evidence for the unique advantages of our particular solar PV segmentation algorithm, that is not the objective of this work. Instead, we have presented evidence demonstrating that practical performance evaluation metrics, which are used to determine how well these techniques will work in practice, are sensitive to a myriad of common pitfalls in the experimental setup and choice of test data. These sensitivities may prevent accurate performance metrics from being obtained, impacting researchers and practitioners alike who wish to apply these techniques in practical settings. These pitfalls may also prevent head-to-head comparison of the many studies in the field on the topic of solar identification, including those of the authors of this work. We summarize each of those findings below along with a recommendation for overcoming each challenge.

**Finding 1**: *Distribution shift may result in significant differences between results*. Differences in the data used for fine-tuning and testing processes, even when the model training is identical and the fine-tuning process is extremely similar, can result in notable differences when the data are different in terms of geography and regional characteristics such as housing density. Moreover, distribution shift in solar PV assessment applications likely applies to many other differences in terms of sensor modality and atmospheric conditions during data collection. When different datasets are used to evaluate performance, there will likely be differences in at least one of these characteristics, making discrepancies in performance metrics likely even between otherwise equivalent algorithms and experimental designs.

This presents two key challenges. (1) This finding implies that unless the test dataset used to validate the data very closely represents the setting in which the technique will be applied in practice (including geography, sensor modality, seasonality, atmospheric condition, etc.), the performance results will likely be biased and will serve as poor representations of future performance. (2) Additionally, if two experiments use two different test datasets for evaluating performance, comparing results between them is of limited meaning.

*Recommendations.* For challenge (1), the trustworthiness of generalization performance estimates, ensuring that the test dataset is as similar as possible to the setting in which the technique will be applied will be key to effective performance estimation. If the algorithm will be applied globally, then the test data should have global coverage. If the algorithm will only be applied to 0.3 meter Worldview 3 satellite data, then the test data should be composed of such data. If the algorithm will be applied to a range of data sources and geographies, then the test data should be as close to that mix as possible.

Challenge (2), comparing performance across studies, is fortunately a much simpler fix. As a community, we should adopt public benchmark datasets with predefined training and test datasets, and at minimum include the results from those benchmarks in every study. Results from other datasets could be used, but including at least a few core benchmark results, the community could unambiguously measure and more quickly advance progress on automated solar PV identification. Sharing easily reproducible code would also enhance innovation and replication. Towards this end, and as we mentioned earlier, we have shared our Connecticut dataset[2] used in this study as well as our codebase[3].

**Finding 2**: *Ground truth is not always trustworthy*. Whether they are data from a records aggregator, manual annotations from human annotators, or regional estimates of solar capacity, the data that we treat as ground truth may be imperfect, even with the best intentions and methodologies of the data providers. If a government agency that captures data on all of the solar connected to the grid is missing systems, it may be simply because the missed systems are not grid-connected or the data entry process for these systems introduced error. Similarly, if we have a dataset on the total municipal capacity of installed solar in 2016, the amounts may vary if the cut-off date on the analysis was January 1 vs December 31 of that year. We found that the data with the fewest inaccuracies for the location of solar PV could be obtained through the manual annotation of satellite imagery. Regardless, the use of imperfect ground truth for evaluation will result in imperfect estimates of performance.

*Recommendations*. All ground truth data used to evaluate performance should be approached cautiously and whenever possible, quality assurance and quality measurement strategies should be used to evaluate the trustworthiness of every dataset. Since manually annotated ground truth was the least error-prone in our study, and we imagine the error rates can be further reduced through the deployment of additional redundant annotators, we recommend using manual annotations as the primary source of ground truth for studies on this topic.

**Finding 3**: *Scale of the analysis impacts the performance assessment*. We presented three levels of scale for performance evaluation of solar PV array segmentation: (1) pixel-level, (2) individual installation (a.k.a. parcel) level, and (3) the regional (in this case municipal) level. The performance metrics that were used in each situation and the values varied since each scale provided a different window into different aspects of performance. These different aspects of performance many be of relevance to practitioners evaluating whether the approach is acceptable for their application. However, if limitations in the scale of the performance are provided, it may cause practitioners who are using those performance metrics to evaluate their decisions to choose incorrectly.

*Recommendations*. Whenever possible, provide performance assessments at multiple scales of aggregation to allow other researchers and practitioners to evaluate the abilities of an algorithm. If limited to one scale, always select the individual array/installation level, since these can often be aggregated up to evaluate performance at other scales, but the reverse is not true. Additionally, providing aggregate estimates only may obscure individual array evaluation quite negatively since it is possible that multiple types of errors cancel out (missing some PV and adding in false positives) to get correct regional estimates, but at the individual installation/parcel level, the estimates could be lacking. Providing the individual/parcel level estimates negates this problem.

In summary, automated solar PV assessment algorithms offer the potential for fast, frequent, and global monitoring of the status of solar PV, even for small-scale solar. With solar PV rapidly increasing as a global energy source, such technologies are ever more relevant to help inform system planning to ensure reliability, economic operation, and expanding access to these technologies. However, as a community, we need to be careful when evaluating the performance of these systems so that we can best inform and assist those practitioners looking to use these data.

[2] Connecticut Solar PV Semantic Segmentation Dataset: https://doi.org/10.6084/m9.figshare.18982199.v5
[3] SolarMapper with MRS (Models for Remote Sensing): https://github.com/energydatalab/solarMapper

## ACKNOWLEDGEMENTS

We thank the NVIDIA corporation for donating a GPU for this work, and the XSEDE and the Duke Compute Clusters for providing computing resources. This work was supported in part by the Alfred P. Sloan Foundation. The content is solely the responsibility of the authors and does not necessarily represent the official views of the Alfred P. Sloan Foundation. The LBNL portion of this work was funded by the U.S. Department of Energy Solar Energy Technologies Office, under Contract No. DE-AC02-05CH11231.

## APPENDIX A: EXPERIMENTAL OVERHEAD IMAGERY DATASETS

### *A. The Duke solar array annotation dataset*

This dataset contains aerial imagery and manually annotated polygons of solar photovoltaic arrays for four cities in California. The United States Geological Survey provides high-resolution aerial orthorectified imagery at 30cm across many metropolitan areas [41]. From these data, 601 images representing four cities in California were selected for inclusion in this dataset (Fresno, Stockton, Modesto, and Oxnard). Selected cities had full imagery coverage from 2013 and a high concentration of solar arrays. For each of these images, the pixel (and geospatial) coordinates of polygon vertices were manually annotated by trained human annotators, resulting in over 19,000 annotations of solar PV arrays. These data are organized into tiles of size 5000×5000 pixels (or 2.25 $km^2$) and the data are summarized in Table A.1 below. Further details about this dataset can be found in [42].

TABLE A.1
SUMMARY OF SELECTED COLOR ORTHOIMAGERY DATA IN CALIFORNIA

| City | Number of image tiles | Area | Number of annotations | Area of annotations |
|---|---|---|---|---|
| Fresno | 412 | 927 $km^2$ | 13803 | 0.473 $km^2$ |
| Stockton | 94 | 211 $km^2$ | 2162 | 0.109 $km^2$ |
| Modesto | 20 | 45 $km^2$ | 609 | 0.051 $km^2$ |

### *B. The Connecticut satellite imagery dataset*

This dataset contains very high (*3 inch, 7.62 cm*) resolution aerial imagery for the entire state of Connecticut. This dataset was provided through the Connecticut Department of Energy and Environmental Protection shared via the University of Connecticut [43]. Collected in 2016, this orthophotography and lidar dataset provides aerial imagery (a) by tile in three formats (GeoTIFF, MrSID3 and MrSID4) and (b) by town mosaic in two formats (MrSID3 and MrSID4). From these data, we extracted all GeoTiff tiles covering the entire state of Connecticut. There are 22,634 tiles, totaling 8.48 Tb of data. Each tile is 2,500ft on a side or 0.224 sq. miles, 143.5 acres and 6,250,000 sq. feet. Each pixel represents approximately 3 inches.

For our work, since it is unlikely that such high resolution imagery will generally be available at a large scale, we downsampled all of the imagery to a resolution of *30 cm*, which is the same resolution as most high resolution satellite imagery (summary statistics showed in Tabel A.2).

TABLE A.2
SUMMARY OF DOWNSAMPLED COLOR ORTHOIMAGERY IN CONNECTICUT

| Dataset | Number of image tiles | Area of image tiles |
|---|---|---|
| Full Connecticut Dataset | 22,634 | 13152 $km^2$ |

### *C. The Duke Connecticut solar imagery dataset*

To develop a training dataset to fine-tune our algorithm and evaluate its performance in the State of Connecticut, we developed a training dataset by manually annotating a subset of the full CT imagery. This subset is comprised of 87 tiles that were manually annotated, producing 1,611 solar PV array annotations. As summarized in Table A.3 as we split the 87 tile subset into a fine-tuning set and a validation set at a 2:1 ratio.

TABLE A.3
SUMMARY OF DUKE ANNOTATED COLOR ORTHOIMAGERY IN CONNECTICUT

| Dataset | Number of image tiles | Area of image tiles | Number of annotations | Area of annotations |
|---|---|---|---|---|
| Fine-tuning set | 57 | 33.12 $km^2$ | 608 | 0.058 $km^2$ |
| Validation set | 30 | 17.43 $km^2$ | 1003 | 0.091 $km^2$ |

### D. The LBNL San Diego satellite imagery dataset

This dataset, provided by Lawrence Berkeley National Laboratory through collaboration, covers selected regions of the city of San Diego, California at a resolution of *30 cm*. We manually annotated 40 tiles of the imagery and split it into a fine-tuning set and a validation set with a 2:1 ratio to demonstrate SolarMapper's capability in another unique geographic region. Statistics of the LBNL San Diego dataset are summarized in Table A.4.

TABLE A.4
SUMMARY OF ANNOTATED COLOR ORTHOIMAGERY IN SAN DIEGO, CA

| Dataset | Number of image tiles | Area of image tiles | Number of annotations | Area of annotations |
|---|---|---|---|---|
| Fine-tuning set | 26 | 14.63 $km^2$ | 2482 | 0.050 $km^2$ |
| Validation set | 14 | 7.88 $km^2$ | 1150 | 0.023 $km^2$ |

## APPENDIX B: SOLARMAPPER DETAILS

In this section, we provide the design details of SolarMapper. SolarMapper is essentially a convolutional neural network (CNN) that has been trained to identify solar arrays, and we present the key components of such an algorithm: (i) the structure of the CNN, (ii) the precise manner in which it was trained to recognize solar panels, and (iii) how it can be applied to map solar panels in new imagery.

### A. The SolarMapper structure: U-net

The SolarMapper is based on the popular U-Net architecture [44] for semantic segmentation (i.e., mapping in our context); however SolarMapper employs only half as many convolutional filters in each layer as the original U-net model. This particular design is motivated by its recent success in the INRIA building labeling competition for semantic segmentation of buildings in overhead imagery, in which it achieved the best overall performance.

### B. Pre-training SolarMapper

The weights of the SolarMapper model were optimized in order to minimize a standard pixel-wise cross entropy loss [45] over the entire training dataset of imagery. The loss for each individual pixel is given by

$$CE = \sum_{i \in C} \lambda \, t \log(y) + (1-t)\log(1-y) \tag{B.a}$$

Where $t \in \{0,1\}$ corresponds to the true class identity of an input pixel (i.e., $t = 1$ if it is a array pixel), and $y \in [0,1]$ is the probability that the pixel is a solar array, as estimated by SolarMapper. The parameter $\lambda$ controls the relative importance of panel pixels and non-panel pixels, respectively. In this work we set $\lambda = 2.3$, to increase the influence of solar array pixels (relative to non-solar-array pixels) on the parameters influenced by the model. This was done to compensate for the relative scarcity of panel pixels in the data.

The sum of the cross-entropy over all pixels in the training dataset is minimized using Stochastic gradient descent [46,47]. For gradient descent we employed mini-batches of 5 image "patches" of size of $512 \times 512$ pixels. With gradient descent we employed the Adam optimizer [48] with learning rate of $10^{-4}$, $\beta_1 = 0.9$, $\beta_2 = 0.999$, and $\epsilon = 10^{-8}$. The learning rate was dropped by a factor of $10^{-1}$ after 50 epochs. No L2 weight regularization was used for our model. SolarMapper was trained for 100 epochs with 8,000 mini-batches per epoch, and each mini-batch contained.

The training dataset was comprised of a random sample of 50% of the image tiles from each of the three cities, respectively, in the Duke Solar Panel Dataset (see Table A.1). This 50% corresponds to "Fold One" of the training data discussed in Appendix C, and the images in it are available in the Supplementary Materials for this paper. Since the image tiles are 5000x5000 pixels, and SolarMapper requires input images of size 512x512 during training, each tile was partitioned into a grid of overlapping 512x512 image "patches". Neighboring patches in the grid overlapped by 92 pixels, corresponding to the amount of imagery at the edges of the input image patches that cannot be processed by the U-net during training (see [44] for details). Subsequently, patches are sampled randomly from the sampling grid. Random rotations of $\theta \in \{0^o, 90^o, 180^o, 270^o\}$ were applied to each input patch.

### *C. Applying SolarMapper to new imagery*

The trained SolarMapper model can be applied to new imagery, assigning a number $p \in [0,1]$ to each pixel indicating the probability that it corresponds to a panel. The fully-trained SolarMapper model is publicly available for download here [49], and is built in the popular PyTorch framework [50].

By default, SolarMapper requires an input image that is 512x512 pixels in size, and returns an output image that is 388x388 pixels in size. This shrinkage is due to zero-padding that occurs within each convolutional layer of the U-net CNN underlying SolarMapper; more details can be found in [44]. Although this is the default input size of the SolarMapper, we note that it is possible to increase the input size of SolarMapper substantially, with a proportional increase in the output size, resulting in better performance and substantially faster processing speeds.

Appendix D provides comprehensive experimental results indicating the performance that can be expected from SolarMapper under ideal conditions. There are two ideal conditions in which to apply SolarMapper: (i) the target imagery is similar to the training imagery in the Duke Solar Panel Dataset, or (ii) it has been sufficiently fine-tuned with hand-labeled examples of panels in the target imagery.

SolarMapper can be directly applied to new imagery, without modification, but it will work best on imagery that is similar to the training imagery. Specifically, the target imagery should be orthorectified aerial photography, captured at ground sampling rate of 0.3m, and collected over structures and geography similar to those in the Duke Solar Panel Dataset.

### *D. Fine-tuning SolarMapper for new imagery*

Although SolarMapper can be directly applied to new imagery without modification, it will work best on imagery that is relatively similar to the training imagery. This implies that, ideally, the target imagery should be orthorectified aerial photography, captured at ground sampling rate of 0.3 m, and collected over structures and geography similar to those in the Duke Solar Panel Dataset. This may be rare however, and therefore it is often necessary to “fine-tune” SolarMapper to work well on the targeted imagery and imaging conditions. As we show in this work, with relatively little additional training imagery, it is possible to fine-tune SolarMapper to substantially different imagery and obtain accurate results.

## APPENDIX C: THRESHOLD FOR RESIDENTIAL AND COMMERCIAL ARRAYS

As we performed the area-based approach to separate residential and commercial PV arrays (Section IV.C), a threshold array size value was used. According to LBNL’s Tracking the Sun report [36], 20 kW is the generation capacity threshold to distinguish residential PV arrays from commercial ones. To convert the generation capacity threshold to an array size threshold, we calculated the generation capacity (kW) and the size ($m^2$) to generation capacity (kW) for over 25,000 solar PV array models documented by NREL’s System Advisor Model [51]. We took the average (4.86 $m^2/kW$), multiplied by the generation capacity threshold of 20 kW, and arrived at the estimated size threshold between residential and commercial solar PV arrays of 97.13 $m^2$. We rounded this threshold value up to 100 $m^2$ in our analysis.

## APPENDIX D: STRATIFIED EVALUATION BY BUILDING DENSITY

The difficulty of identifying solar PV from overhead imagery could vary substantially by location due to the different background of the imagery. To investigate how the building density in the surrounding areas might affect SolarMapper’s capability of identifying solar PV, we retrieved building density information from Microsoft’s US Building Footprints [37] dataset to calculate buildings per image tile. We split all 30 image tiles in the validation set to three stratified brackets by the 25$^{th}$, the 50$^{th}$, and the 75$^{th}$ percentiles (Section IV.D).

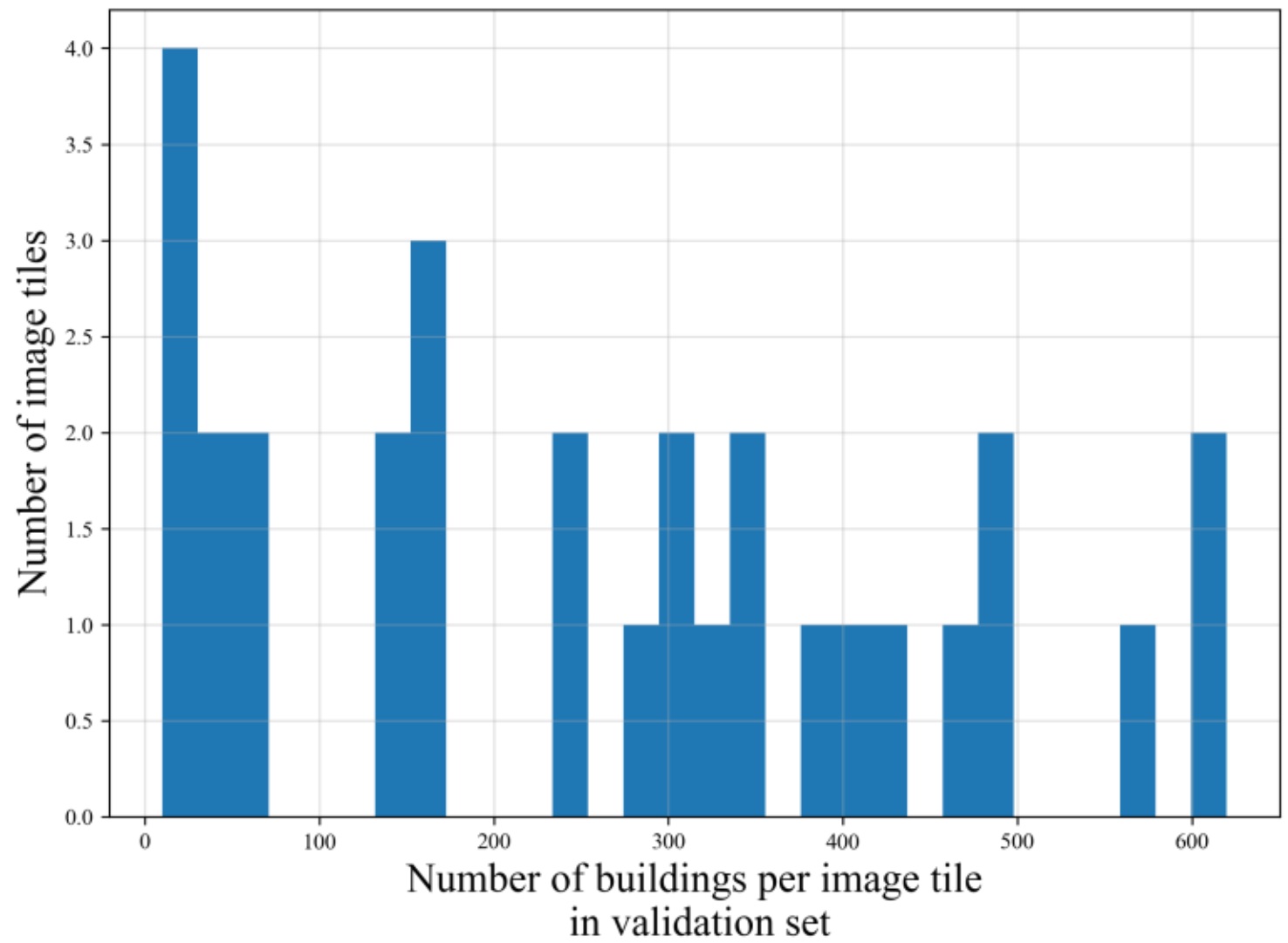

**Fig. D.1: Number of buildings per image tile distribution of the 30 image tiles in the validation set.**

We categorized any tile containing a number of buildings no larger than the 25th percentile (87 buildings in an image tile) in the distribution shown in Fig. D.1 as "Low" building density, any tile containing a number of buildings higher than the 75th percentile (402 buildings in an image tile) as "High" building density, and everything in between as "Medium" building density. Some examples of such images from each of the 3 brackets are shown in Fig. D.2. We then performed array-wise evaluation for each bracket, as well as all validation image tiles as described in Section V.A.

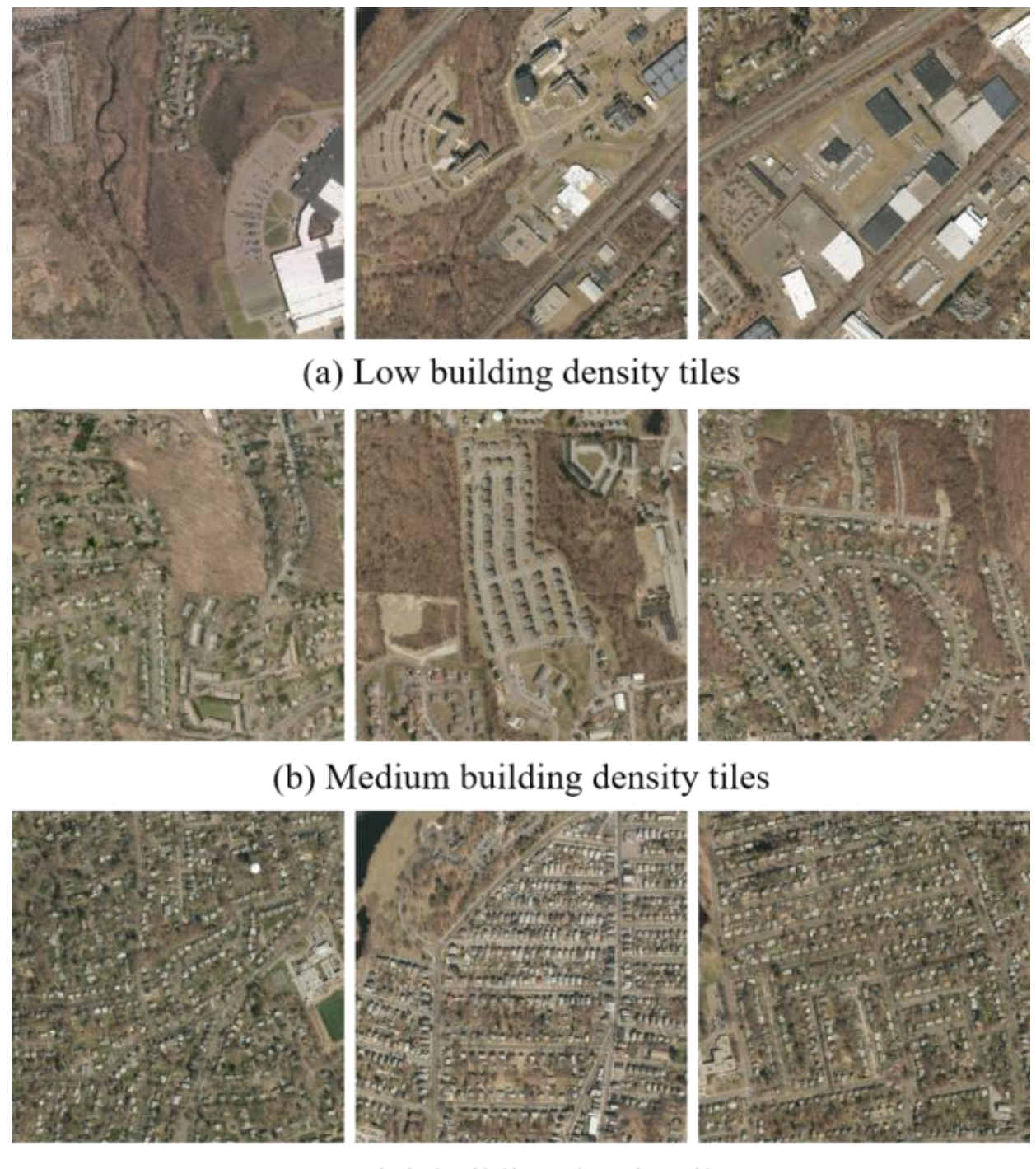

(a) Low building density tiles

(b) Medium building density tiles

(c) High building density tiles

**Fig. D.2: Example image tiles from the 3 building density brackets.**

## APPENDIX E: PANEL CAPACITY ESTIMATION

### *A. The Connecticut Solar Scorecard dataset*

This dataset from Data Driven Yale, let by Angel Hsu, provides an estimate of installed residential solar PV capacity by municipality for all of Connecticut [52]. While the original purpose of the data was to assess the contribution of individual municipalities to the Connecticut state goal for 300MW of residential solar power by 2020, this dataset produced scorecards ranking municipalities in five categories.
The 169 CT municipalities are assigned a score for each of the following:

- **Solar adoption**: the amount of residential solar energy installed
- **Clean energy engagement**: the degree of financial support of solar PV deployment by municipalities
- **Information availability**: efforts municipalities make to popularize solar energy
- **Permit process**: ease of the permit acquisition process
- **Time & cost**: permit submission time, permit turnaround time, permit fee.

In addition to the above scoring metrics, which are further aggregated to produce an overall score, the dataset includes among others, the following information for each municipality:

- Residential solar capacity (kW) at the beginning of 2016
- Residential solar capacity (kW) in April 2016
- Number of PV arrays per 1000 homes
- Average permit fee
- Fraction of electricity from solar by municipality

For our purposes, the residential solar capacity in kW is the quantity we use to evaluate our algorithms performance in estimating the total installed solar PV. Of course, this is focused on residential; therefore in our analysis we classify PV arrays as being either residential or commercial and use this dataset only for comparing residential capacity.

*B. Groton as an outlier*

We discovered that, despite Groton's official capacity estimates being among the lowest in the state of CT, it does indeed have a large number of residential solar arrays. Upon manual inspection, we discovered a large residential community in which nearly every roof contains a solar array (Fig E.1). We hypothesize that this was not a typical residential project and should be excluded from the regression analysis using the CT Solar Scorecard data as the ground truth.

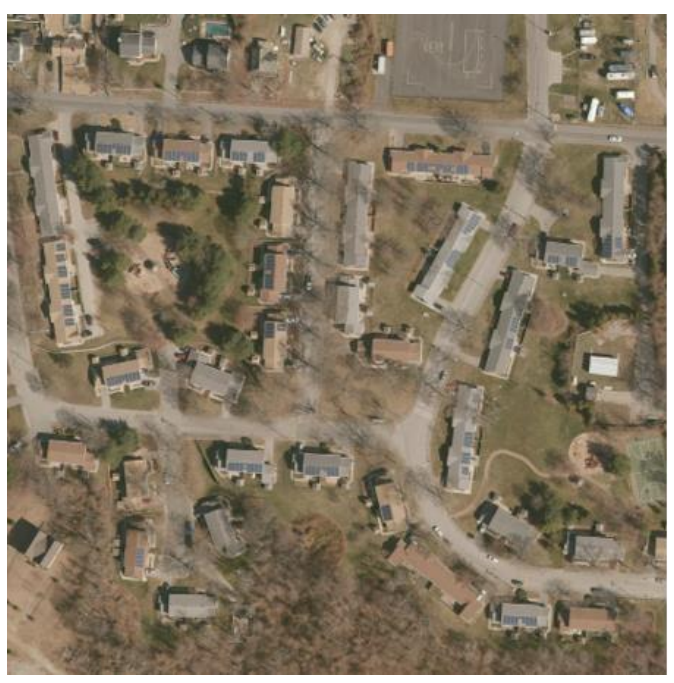 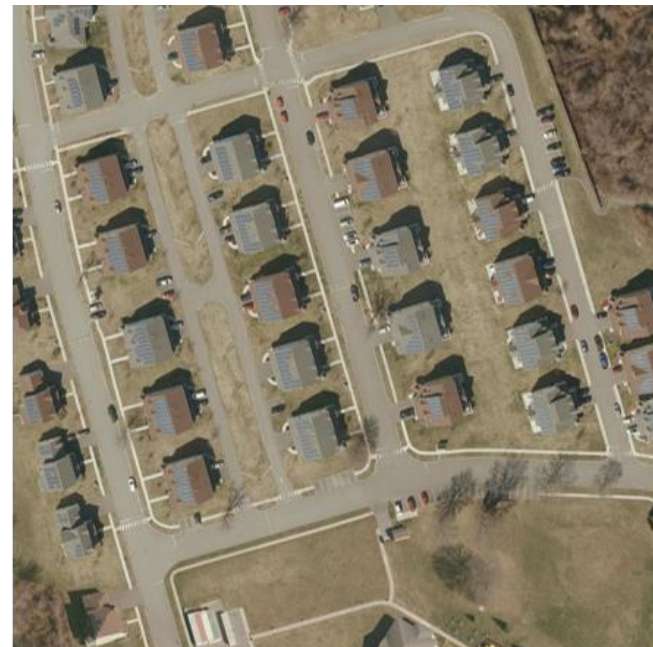 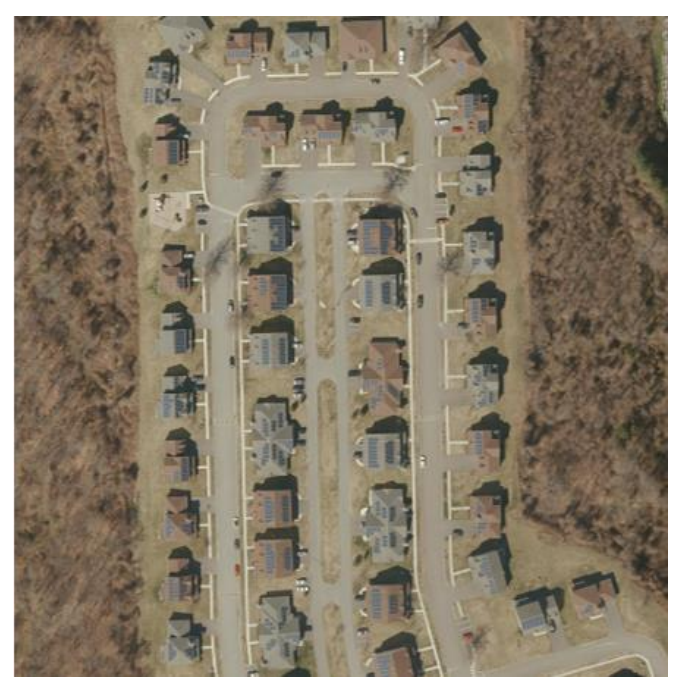

**Fig. E.1: Examples: Residential community in Groton, CT where nearly every roof contains a solar array.**